\documentclass{article}
\usepackage{amsmath,epsfig}
\usepackage[preprint]{spconfa4}

\usepackage{graphicx}
\usepackage{amssymb}

\usepackage[noend]{algpseudocode}
\usepackage{algorithmicx,algorithm}
\usepackage{booktabs}
\usepackage{multirow}
\usepackage{graphbox}
\usepackage{array,booktabs}
\usepackage{amsmath,amssymb}
\usepackage{soul}
\usepackage{xcolor}
\usepackage{bbding}
\usepackage[misc]{ifsym}

\newcommand{\eg}{\emph{e.g.}}
\newcommand{\etal}{\emph{et al.}}
\newcommand{\ie}{\emph{i.e.}}

\let\OLDthebibliography\thebibliography
\renewcommand\thebibliography[1]{
  \OLDthebibliography{#1}
  \setlength{\parskip}{0pt}
  \setlength{\itemsep}{0pt plus 0.3ex}
}

\begin{document}\sloppy

\def\x{{\mathbf x}}
\def\L{{\cal L}}

\title{MaskGroup: Hierarchical Point Grouping and Masking for 3D Instance Segmentation}
\name{Min Zhong$^{1,2}$, Xinghao Chen$^2$, Xiaokang Chen$^1$, Gang Zeng$^1$, Yunhe Wang$^2$} %
\address{$^1$ Key Laboratory on Machine Perception, Peking University\\
	$^2$ Huawei Noah's Ark Lab\\
	\small\{xinghao.chen, yunhe.wang\}@huawei.com, \{minzhong, zeng\}@pku.edu.cn
}

\maketitle

\begin{abstract}
This paper studies the 3D instance segmentation problem, which has a variety of real-world applications such as robotics and augmented reality. Since the surroundings of 3D objects are of high complexity, the separating of different objects is very difficult. To address this challenging problem, we propose a novel framework to group and refine the 3D instances. In practice, we first learn an offset vector for each point and shift it to its predicted instance center. To better group these points, we propose a Hierarchical Point Grouping algorithm to merge the centrally aggregated points progressively. All points are grouped into small clusters, which further gradually undergo another clustering procedure to merge into larger groups. These multi-scale groups are exploited for instance prediction, which is beneficial for predicting instances with different scales. In addition, a novel MaskScoreNet is developed to produce binary point masks of these groups for further refining the segmentation results. Extensive experiments conducted on the ScanNetV2 and S3DIS benchmarks demonstrate the effectiveness of the proposed method. For instance, our MaskGroup achieves a 66.4\% mAP with the 0.5 IoU threshold on the ScanNetV2 test set, which is 1.9\% higher than the state-of-the-art method.
\end{abstract}
\begin{keywords}
Point Cloud, 3D Instance Segmentation, Hierarchical Point Grouping
\end{keywords}

\section{Introduction}

\begin{figure}[t!]
	\begin{center}
		\includegraphics[width=0.7\linewidth]{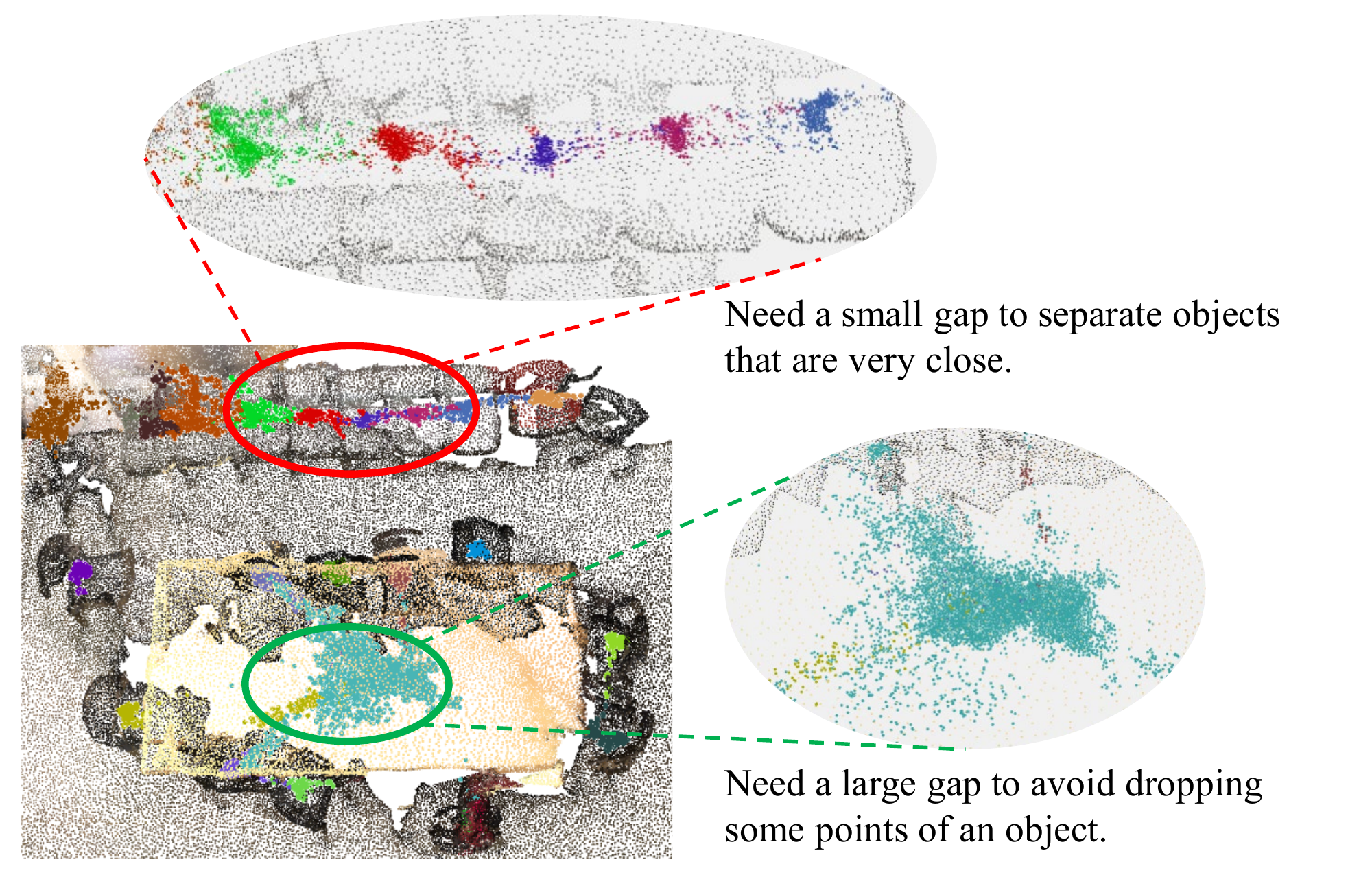}
	\end{center}
	\vspace{-7mm}
	\caption{Illustration of 3D points in complex environment.}
	\label{fig:intro}
	\vspace{-6mm}
\end{figure}

To tackle the 3D instance segmentation problem,
a series of detection-based  methods~\cite{3dbonet, 3dsis} are explored for predicting 3D bounding boxes from the observed point data. These methods produce a mask to obtain the instance inside the bounding box.
Moreover, the embedding-based approaches~\cite{jsis3d,asis,3dmpa,pointgroup,occseg} learn a spatial or feature embedding vector for each point, and utilize the clustering algorithm to obtain the instance.
For example, Jiang~\etal~\cite{pointgroup} proposed PointGroup, an end-to-end bottom-up architecture which groups the 3D points by considering the void space between objects. PointGroup first learns a space offset to move the object points towards their instance centers.
Then, all points are clustered according to the distance of the blank space. Two points within a certain distance are merged into one group. 
However, it is hard to decide on a specific single distance to meet a variety of situations, because objects are lying from each other with different distances (As shown in Fig.~\ref{fig:intro}). Additionally, since the objects in the 3D space are of high complexity, noisy points are usually existing in the grouped instances, which hinders the performance.

To enhance the performance of 3D instance segmentation, we present a masked hierarchical point grouping approach, namely MaskGroup. In practice, we first learn the offsets and semantic labels for the input 3D points. Since the sizes of observed 3D objects in the given environment are often various, we then propose a Hierarchical Point Grouping (HPG) algorithm for carefully merging the points of each object progressively. Specifically, a small gap is utilized to cluster points into several initial groups. Then we exploit progressively enlarged distances to merge small groups from the previous step into larger groups. The clustered groups obtained in different steps contain multi-scale information which can be further used for the final instance predictions with better results.

In addition, the clustered points obtained using the hierarchical grouping may have some noisy points for each instance due to the complex enviroment of 3D point data. To effectively refine and evaluate these groups, we propose a tiny MaskScoreNet to mask out the background points via a mask branch and simultaneously evaluate the quality of each mask via a score branch. The mask branch predicts a binary mask for the points in a group to distinguish the actual points in object instances. The score branch further predicts a binary quality score for the masked group.

Our contributions are summarized as follows: (1) In order to make full use of the multi-scale information of 3D instances, we propose a Hierarchical Point Grouping algorithm to merge the points into different groups progressively. (2) We proposed a novel  MaskScoreNet to refine the clustered groups, which produces binary point masks for all grouped instances to eliminate noisy points and predicts confidence scores for final instances, simultaneously. (3) The proposed method, \ie, MaskGroup, achieves state-of-the-art results on the challenging ScanNetV2 and S3DIS benchmarks, demonstrating its effectiveness for 3D instance segmentation.

\begin{figure*}
	\begin{center}
		\includegraphics[width=0.73\linewidth]{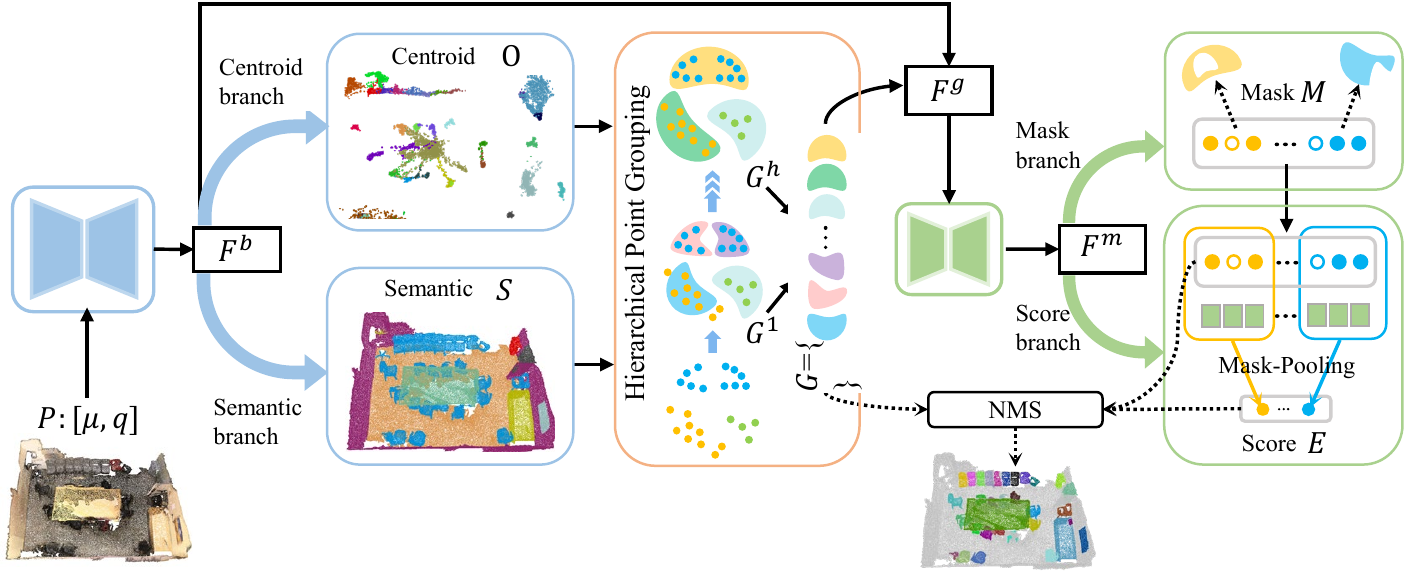}
	\end{center}
	\vspace{-4mm}
	\caption{The overall architecture of our proposed MaskGroup.}
	\label{fig:overview}
	\vspace{-5mm}
\end{figure*}

\section{Related Work}

3D instance segmentation for large scale point clouds has attracted great research interest~\cite{liang2021instance,chen2021hierarchical,pointgroup,occseg,3dmpa} in recent years.
Similar to the detection-based 2D instance segmentation methods, Kundu~\etal~\cite{3drcnn} proposed 3D-RCNN, which builds upon Faster R-CNN  to predict 3D bounding boxes from point clouds.
Yang~\etal~ proposed BoNet~\cite{3dbonet}, which predicts the box from the global point features and produces a point mask for the bounding box.

The embedding-based idea in 2D instance segmentation is also extended into the 3D domain.
Wang~\etal~\cite{sgpn} proposed to learn the neighborhood relationship for each pair of points and access the instance according to similar relations.
Several methods~\cite{jsis3d,asis} extended the 2D embedding idea to point cloud, which encourage points in the same instance to lie close in the embedding space, and adopt mean-shift algorithm for clustering.
Qi~\etal~proposed VoteNet~\cite{votenet}, which learns shift vectors to move points towards their instance centers and can be seen as a kind of spatial embedding.
Engelmann~\etal~\cite{3dmpa} proposed to learn a spatial embedding of points and group points within a certain distance into one group, and use learned feature embedding to merge the group.
Jiang~\etal~\cite{pointgroup} proposed PointGroup, which also learns the spatial embedding, but groups the points connected with each other into an instance.
OccuSeg~\cite{occseg} utilizes both spatial and feature embedding to progressively group the point into an instance.

{The hierarchical approach is also used in 3D scene analysis~\cite{shi2019hierarchy,armeni20193d}, which is mainly utilized for bottom-up context aggregation for hierarchical scene graph.  Our paper proposes the hierarchical grouping and masking to capture multi-scale information for instance prediction, which is quite different from the above mentioned papers.}
\section{The Proposed MaskGroup}%

\subsection{Method Overview}

3D instance segmentation task aims to separate different objects in 3D scene and predict the semantic label of each instance. Our proposed method is depicted in Fig.~\ref{fig:overview}.

The input 3D scene is represented as a point cloud $\boldsymbol{P}$ with $N$ points in total, \ie, $\boldsymbol{P} = \{p_i\}_{i=0}^{N} \in \mathbb{R}^{N\times k_0} $, where $k_0$ is the  channel number of point feature such as the point location  $\mu_i = (x_i, y_i, z_i)$ and color $q_i = (r_i, g_i, b_i)$. 
Then a voxel-based backbone network~\cite{sparse3d} is applied to extract the 3D features $\boldsymbol{F^b}=\{F^b_{i}\}_{i=0}^{N} \in \mathbb{R}^{N \times k_1}$, where each point $i$ has a corresponding feature vector $F^b_{i} \in \mathbb{R}^{k_1}$,  $k_1$ is the feature length.
After that, two sub-branches are exploited to obtain the semantic labels $\boldsymbol{S}=\{s_i\}_{i=0}^{N} \in \mathbb{R}^{N}$ and the offset vectors $\boldsymbol{D}=\{d_i\}_{i=0}^{N} \in \mathbb{R}^{N\times 3}$, where $d_i = (\Delta x_i, \Delta y_i, \Delta z_i)$. 
Adding the offsets to original coordinates, we obtain the estimated object centroids $\boldsymbol{O}=\{o_i\}_{i=0}^{N} \in \mathbb{R}^{N\times 3} $, where $o_i = \mathbf{\mu}_i + d_i$.

In order to separate objects with the same semantics, we explore the void space in $\boldsymbol{O}$ or $\boldsymbol{P}$.
Given a clustering radius $r_1$ manually, we can merge the points within $r_1$ into $|G^1|$ groups $G^1 = \{G^1_i\}_{i=0}^{|G^1|}$.
However, it is difficult to select a proper $r_1$ that works well for instances with different scales. 
To this end, we proposed a hierarchical point grouping algorithm to cluster the points via a multi-scale scheme. 
First, a small radius is utilized to cluster points into several small groups. Then we exploit progressively enlarged clustering radius to merge small groups into larger groups for  $H$ steps. 
The clustered groups $\boldsymbol{G} = \{G^1 \cup ... \cup G^H\} = \{G_i\}_{i=0}^{|G|}$ obtained in different steps contain multi-scale information and are further used for final instance prediction.

However, the roughly clustered groups $\boldsymbol{G}$ may contain some noisy points.
We then propose a MaskScoreNet to refine and evaluate these groups.
For each group $G_i$, MaskScoreNet predicts a binary mask $M_{i}$ to eliminate noisy points in this group and also outputs a score $E_{i}$ to indicate the confidence score of this masked group.

In inference stage, the groups $\boldsymbol{G}=\{G_i\}_{i=0}^{|G|}$, masks $\boldsymbol{M}= \{M_i\}_{i=0}^{|M|}$ and their corresponding scores $\boldsymbol{E}= \{E_i\}_{i=0}^{|E|}$ ( $|G|=|M|=|E|$) are fed into the Non Maximum Suppression (NMS) module to obtain final instance predictions.

\subsection{Backbone Network Architecture}

The backbone network consists of a feature extraction network and two sub-branch networks for semantic prediction and object centroid regression
The feature extraction network takes points $\boldsymbol{P}$ as input and obtains point features $\boldsymbol{F^b}$.

A multi-layer perceptron (MLP) is applied to produce semantic scores $\boldsymbol{B} = \{b_i\}_{i=0}^{N} \in \mathbb{R}^{N \times {C}}$ for the $N$ points, where $C$ is the number of semantic classes.
The semantic labels $\boldsymbol{S} = \{s_i\}_{i=0}^{N} \in \mathbb{R}^{N}$  are the classes with the highest score.
This semantic branch is supervised by a semantic segmentation loss $\mathcal{L}_{\mathrm{sem}} = \frac{1}{N}\sum_{i=1}^{N} \mathcal{H}( b_i, \hat{s}_i )$, 
where $\hat{s}_i$ is the ground truth semantic label of point $i$, $\mathcal{H}(\cdot)$ represents the cross entropy function.

Another MLP encodes $\boldsymbol{F^b}$ to produce 3-dimensional offset vectors $\boldsymbol{D}$ for the $N$ points.
This centroid branch is supervised using the following  $L_1$ regression loss:
\begin{equation}
	\small
	\mathcal{L}_{\mathrm{off}} =  \frac{1}{| \hat{I} |}  \sum_{\hat{i}=1}^{| \hat{I} |}  \frac{1}{N_{\hat{i}}}  \sum_{i=1}^{N_{\hat{i}}}||{d}_i + \mathbf{\mu}_i- \hat{o}_i||,
\end{equation}
where $\hat{o}_i$ is the ground truth centroid of $i$-th point that belongs to $\hat{i}$-th instance, $N_{\hat{i}}$ is the number of points in the $\hat{i}$-th instance and $|\hat{I}|$ is the number of ground truth instance.

To ensure that the points are moving towards their instance centroids, following~\cite{mtml}, we adopt a direction loss $\mathcal{L}_{dir}$ to constrain the direction of predicted offset vectors:
\begin{equation}
	\small
	\mathcal{L}_{\mathrm{dir}} = -\frac{1}{| \hat{I} |}  \sum_{\hat{i}=1}^{| \hat{I} |}  \frac{1}{N_{\hat{i}}}  \sum_{i=1}^{N_{\hat{i}}}  \frac{d_i} {||d_i||_2} \cdot \frac{\hat{o}_i - \mathbf{\mu}_i }{||\hat{o}_i - \mathbf{\mu}_i||_2},
\end{equation}
where $\mathbf{\mu}_i$ represents the 3D position of the $i$-th point of the $\hat{i}$-th instance. 
With the learned offset vector $d_i$, we can obtain the estimated object centroid $o_i = \mathbf{\mu}_i + d_i$.

\subsection{Hierarchical Point Grouping}

With spatial information like 3D point positions and estimated object centroids, we can cluster points into different groups.
Traditional methods for instance clustering in 3D instance segmentation rarely consider the multi-scale information, \eg, PointGroup~\cite{pointgroup} utilized a single radius threshold for point grouping, which is insufficient for capturing information for instances with different scales. Multi-scale information has been widely used in 2D object detection and instance segmentation~\cite{fpn,chen2021empirical}, \eg, Feature Pyramid Networks~\cite{fpn}. 
And some hierarchically ideas are used for analyzing 3D scenes~\cite{DBLP:conf/iccv/ArmeniHZGMFS19, DBLP:conf/cvpr/ShiCWS019}.
However, it is still less explored for 3D instance segmentation.

To this end, we propose a hierarchical point grouping algorithm that takes multiple spacing distance into consideration.
Specifically, we take the groups $G^{h-1}$ found in $(h-1)^{th}$ round as input. 
Especially, when $h=1$, each group in $G^{0}$ is a single point.
In the $1^{st}$ round, we use a small spatial distance $r_1$ to group the points.
If two points have the same semantics and the distance between them is smaller than $r_1$, then we put them into the same group.
At the end of the $1^{st}$ round, we obtain a set of initial groups $G^1=\{G^1_i\}_{i=0}^{|G^1|}$ with a number of $|G^1|$.

In the next round, we take groups $G^1$ in the first round as inputs.
For two groups $G^1_i$ and $G^1_j$ with the same semantics, we merge them into a new group if the distance between two groups is smaller than radius $r_2$. 
All point within a group has the same semantics, so the semantic label of a group $G_i$ is defined as the label of its point,
We denote the semantic label of group $G_i$ as $\boldsymbol{S}[G_i]$.
The distance  $||G_i,G_j||$ between group $G_i$ and $G_j$ is calculated as the minimum distance between points in these two groups.
In this way, we obtain some new groups  $G^2=\{G^2_i\}_{i=0}^{|G^2|}$ with a number of $|G^2|$.

We repeat this process for $H$ times. For each round, we have $r_{h-1} < r_{h}$, so we can merge groups in last round into larger groups gradually. 
We collect the groups in all rounds and eliminate the groups in which the number of its points is fewer than $N_{\theta}$, 
finally get  $\boldsymbol{G} = \{G^1 \cup ... \cup G^H\}$.
These point groups obtained in different stages contain clustering information for different scales. For example, since $G^1$ is clustered with a small radius, point group in $G^1$ can better segment instances with small scales. 
Meanwhile, larger instances are better segmented in $G^H$. Point groups with multiple scales $\boldsymbol{G}$ will be exploited for final instance prediction, which is beneficial for more accurate 3D instance segmentation.

\subsection{MaskScoreNet}

After the hierarchical point grouping step, we obtain a set of groups $\boldsymbol{G} = \{G_i\}_{i=0}^{|G|}$, where $|G|$ is the total number of groups collected from all grouping steps.
Group  ${G_i}$ consists of $N^i_g$ points, \ie, $G = \{p_j\}_{j=0}^{N_g}$ (we omit $i$ for simplicity).
Some groups in $\boldsymbol{G}$ are well grouped and some contain inaccurate points. Therefore, we design a novel MaskScoreNet to refine the groups $\boldsymbol{G}$ and also evaluate the quality of each group.

Recall that, each point $i$ has a corresponding backbone feature vector $F^b_{i} \in \mathbb{R}^{k_1}$.
So, for a group  $G = \{p_j\}_{j=0}^{N_g}$, we can create a group feature ${F^g} \in \mathbb{R}^{Ng \times k_1}$ by selecting the corresponding point features from $\boldsymbol{F^b}$,

The group ${G}$ assigned with group feature ${F^g}$ is fed into a small U-Net  to better aggregate the group information, the output features ${F}^{m}$ are fed into a MLP (mask branch) to predict a binary mask $M = \{m_j\}_{j=0}^{N_g}$ (we omit $i$ for simplicity), 
where $m_j$ is the confidence score for $j$-th point in $G$, which indicates the probability of this point belonging to $G$. 
The mask loss function for all groups is formulated as:
\begin{equation}
	\small
	\resizebox{0.91\hsize}{!}{
		$\mathcal{L}_{mask} = - \frac{1}{|G|} \sum_{g=0}^{|G|}  \frac{1}{N_g} \sum_{i=0}^{N_g} (\hat{m}_i log(m_i) + \\
		(1 - \hat{m}_i) log(1 - m_i)),$
	}
\end{equation}
where $\hat{m}_i$ is the ground-truth mask value corresponding to $m_i$.
To obtain the ground-truth mask, we first found out the ground-truth instance $I_{\hat{g}}$ which has the largest Intersection over Union (IoU) between cluster $G$ :
\begin{equation}\small
	\resizebox{0.51\hsize}{!}{$
	\hat{g} = \text{argmax}_i\left(\left\{\text{IoU}(G, I_i) \mid  I_i \in \mathbf{\hat I}\right\}\right),
	$}
\end{equation}
where $\boldsymbol {\hat I}$ is a set of ground-truth instances.
Then we can get the value of $\hat{m}_i$:
\begin{equation}\small
	\hat{m}_i  = 
	\begin{cases}
		1,     & {if~point~i~in~ both ~ {G} ~and ~I_{\hat{g}}}\\
		0,    & {otherwise}\\
	\end{cases}.
\end{equation}

To evaluate the quality of the masked groups, we propose a MaskPooling layer that uses the predicted mask $M$ to average pool the group feature $G$ across the points, and obtain a feature vector $F^e$ to represent the masked group.
After that, a  score branch (MLP and Sigmoid) is utilized to predict the quality score ${E}$ for the masked group based on $F^e$.
The score loss is defined as:
\begin{equation}
	\small
	\mathcal{L}_{score} = - \frac{1}{|G|} \sum_{i=1}^{|G|} (\hat{E}_i log(E_i) + (1 - \hat{E}_i) log(1 - E_i)),
\end{equation}
where $\hat{E}_i$ is the ground-truth score between $0$ and $1$ for $M_i$, which is decided by the IoU between $M_i$ and its corresponding ground-truth instances~\cite{jiang2018acquisition, li2019gs3d}.

\subsection{Network Training}

We train the whole framework in an end-to-end manner with the total loss as:
\begin{equation}\small
	\mathcal{L} = \mathcal{L}_{sem} + \mathcal{L}_{off} + \mathcal{L}_{dir} + \mathcal{L}_{mask} + \mathcal{L}_{score}.
\end{equation}

\begin{table}
	\centering
	\small
	\caption{Quantitative results on the ScanNetV2 \textit{testing} set in terms of AP, AP$_{50}$ andAP$_{25}$.}	\begin{tabular}{l|c|c|c|c}
		\toprule
		Methods & Publication & AP  & AP$_{50}$  & AP$_{25}$ \\
		\midrule
		3D-SIS~\cite{3dsis} & CVPR'19 &16.1 &38.2 & 55.8 \\
		SALoss~\cite{saloss} & IROS'20 &26.2  &45.9 &69.5 \\
		PanopticFusion~\cite{pfusion} & IROS'19 &21.4 &47.8 & 69.3 \\
		3D-BoNet~\cite{3dbonet} & NeurIPS'19 & 25.3 & 48.8 & 68.7 \\
		3D-MPA~\cite{3dmpa} & CVPR'20 & 35.5 &	61.1	&	73.7	\\
		OccuSeg~\cite{occseg} & CVPR'20 & \textbf{44.3} & {63.4} & {73.9} \\
		PointGroup~\cite{pointgroup} & CVPR'20 & {40.7} & {63.6} &{77.8} \\
		GICN~\cite{gicn} & arXiv'20 & {34.1} & {63.8} & {78.8} \\
		{DyCo3D}~\cite{dyco} & CVPR'21 &39.5 &{64.1} &{76.1}\\
		PE~\cite{pe} & CVPR'21 &39.6 &64.5 &77.6 \\
		\hline
		\bf {MaskGroup (Ours)} & -  & {43.4} & \textbf{66.4} & \textbf{79.2} \\
		\bottomrule
	\end{tabular}
	\label{tab:performance_scannet}
	\vspace{-6mm}
\end{table}

\begin{table}[tb]
	\begin{center}
		\small
		\setlength{\tabcolsep}{1.4mm}
		\caption{{Comparisons with state-of-the-arts on S3DIS}~\cite{s3dis}.}		%
			\begin{tabular}{ l  |c c c | c c c}
				\toprule
				&  & Area5 &  & & 6-Fold &  \\
				Methods  & AP$_{50}$ & mPrec & mRec  & AP$_{50}$ & mPrec & mRec \\
				\midrule
				GICN~\cite{gicn}  & - & 61.5 & 43.2 & - & 68.5 & 50.8 \\
				PointGroup~\cite{pointgroup}  &{57.8} & {61.9} & {62.1} &{64.0} &{69.6} & {69.2} \\
				OccuSeg~\cite{occseg}  &- &- &- &- &\textbf{72.8} &60.3 \\
				MPNet~\cite{he2020learning} &- &62.5 &49.0 &- &68.4 &53.7 \\			
				InsEmb~\cite{he2020instance}  &- &61.3 &48.5 &- &67.2 &51.8\\				
				DyCo3D~\cite{dyco} &- &\textbf{64.3} &64.2 &-&-&- \\
				ICM-3D~\cite{chu2021icm} &- &57.4 &45.0 &-&65.9&49.8 \\
				\hline
				\bf {MaskGroup}  & \textbf{65.0} &{62.9} &\textbf{64.7} &\textbf{69.9} &66.6 &\textbf{69.6} \\
				\bottomrule
			\end{tabular}
		\label{tab:s3dis-compare}
	\end{center}
	\vspace{-8mm}
\end{table}

\section{Experiments}

\subsection{Experimental Setting}

We evaluate the proposed method on two large scale 3D indoor scene datasets, \ie, ScanNetV2~\cite{scannet} and S3DIS~\cite{s3dis}.
ScanNetV2 contains 1613 scenes with 18 object categories, in which all points are annotated with semantic and instance labels.
S3DIS~\cite{s3dis} contains 6 sub-datasets and has 271 scenes in total. All points are annotated with instance labels and one of the 13 semantic labels. 
In the data processing part, we set the voxel size to 0.02m. 
We randomly crop the scene to make sure the number of points in each scene to be equal or fewer than 250k.
In the inference phase, all scenes are fed into the network without cropping.
In the hierarchical point grouping part, we set the clustering radii $r$ as 0.01m, 0.03m, and 0.05m in different stages. {We use the same cluster radius setting for ScanNetV2 and S3DIS datasets. We have tried to use a different combinations of clustering radii for S3DIS, but observe no performance improvement. It demonstrate that our method has good generalization performance on different datasets.}  The minimum cluster point number $N_\theta$ is empirically set as 50. {The threshold of IoU for NMS is set to 0.7}
The proposed method is trained via Adam optimizer with a base learning rate of 0.001.

\subsection{Comparisons with state-of-the-arts}
\textbf{Instance Segmentation on ScanNetV2~\cite{scannet}}. We compare our MaskGroup on the testing set of ScanNetV2 with a number of prior methods, including 3D-BoNet~\cite{3dbonet}, 3D-MPA~\cite{3dmpa}, PointGroup~\cite{pointgroup}, GICN~\cite{gicn}, OccuSeg~\cite{occseg}, Dyco3D~\cite{dyco} and PE~\cite{pe}. We report the results of AP$_{50}$, AP$_{25}$, and AP of different models in Table~\ref{tab:performance_scannet}. Our MaskGroup achieves the highest AP$_{50}$ score of 66.4\% and outperforms all prior methods. 
Specifically, the AP$_{50}$ score of our method is 2.6\% and 2.3\% higher than GICN~\cite{gicn} and DyCo3D~\cite{dyco}, respectively.
Moreover, our MaskGroup obtains 1.9\% higher AP$_{50}$ than the former best solution PE~\cite{pe} which has 64.5\% AP$_{50}$ score. 
Our method has a higher AP score than most of the prior models and obtains slightly worse AP than OccuSeg~\cite{occseg}. However, MaskGroup achieves 3.0\% and 5.3\% better AP$_{50}$ and AP$_{25}$ than OccuSeg, respectively, demonstrating the superiority of our proposed method. 
Table 1 in supplementary material also lists detailed performance on each category and totally our MaskGroup ranks the $1^{st}$ place in 10 out of 18 classes. 
Fig.1 in supplementary material shows the qualitative results on the validation set of ScanNetV2.
We also provide some qualitative results of PointGroup and our proposed MaskGroup, as shown in Fig.2 in supplementary material. Our method obtains better results than PointGroup.

{
\noindent\textbf{Instance Segmentation on S3DIS~\cite{s3dis}}. 
We report the performance of mean precision (mPrec) and mean recall (mRec) on the 5th sub-datasets results, as well as 6-fold cross validation results. As is shown in Table~\ref{tab:s3dis-compare}, our MaskGroup achieves competitive performance with prior methods under different evaluation protocols. Specifically, MaskGroup gets 65.0\% and 69.9\% on AP$_{50}$ in terms of Area5 and 6-Fold results, which are 7.2\% and 5.9\% higher than PointGroup~\cite{pointgroup}, respectively. Moreover, the proposed MaskGroup outperforms recent methods like MPNet~\cite{he2020learning}, InsEmb~\cite{he2020instance} and ICM-3D~\cite{chu2021icm} in terms of mean precision and recall. MaskGroup achieves competitive results with DyCo3D~\cite{dyco} and OccuSeg~\cite{occseg}, with higher recall and slightly lower precision.
}

\begin{table}[t]
	\begin{center}
		\caption{{Ablation results using different modules on the ScanNet v2 validation set.}}
		\small
		\scalebox{0.83}[0.83]{
			\begin{tabular}{l|cccc|cccc}
				\toprule
				&Score   & HPG &Mask &MaskPool& AP & AP$_{50}$ & AP$_{25}$ \\
				\midrule
				A&\checkmark& & & &39.9 &58.9 &70.3   \\	
				\hline		
				B&\checkmark&\checkmark & & &39.2 &58.3 &71.5  	\\	
				C&\checkmark & &\checkmark &\checkmark &40.4 &59.8 &70.5  \\
				D&\checkmark &\checkmark  &\checkmark & & 39.4 &61.1 &72.7	\\	
				\midrule
				\bf E&\checkmark&\checkmark &\checkmark &\checkmark & \bf 41.9 &\bf 62.7 &\bf 73.6  \\
				\bottomrule
			\end{tabular}
		}
		\label{tab:ab}
	\end{center}
	\vspace{-7mm}
\end{table}

\begin{table}[t]
	\small
	\centering
	\caption{Ablation results using different grouping radii.}
	\scalebox{0.9}[0.9]{
		\begin{tabular}{l|c|c|c}
			\toprule
			Hierarchical Grouping  & AP  & AP$_{50}$  & AP$_{25}$  \\
			\midrule
			\{0.01\}  &37.5 &59.4 &71.0 \\
			\{0.03\}  &41.5 &61.9 &72.3 \\
			\{0.05\}  &41.4 &61.3 &72.0 \\
			\midrule
			\{0.01, 0.03\}  &41.5 &62.6 &72.9 \\
			\bf \{0.01, 0.03, 0.05\}  & 41.9 & 62.7 &  73.6 \\
			\{0.01, 0.03, 0.05, 0.07\} &42.1  &63.0   &73.8 \\
			\{0.01, 0.03, 0.05, 0.07, 0.09\} &42.2 &63.3  &74.0 \\
			\{0.01, 0.03, 0.05, 0.07, 0.09, 0.11\} &42.0  &63.3  &74.2 \\
			\bottomrule
		\end{tabular}
	}
	\label{tab:ab_hgp}
	\vspace{-4mm}
\end{table}

\subsection{Ablation Studies}
\label{sec_exp}

In this section, we conduct extensive ablation studies on the ScanNetV2 validation set to analyze the impacts of different modules in MaskGroup. 

\noindent\textbf{Without Mask Prediction.}
We remove the mask branch and  MaskPool layer from our MaskGroup model to examine the effectiveness of MaskScoreNet.
This degraded model B only obtains the AP score of 39.2\%,  AP$_{50}$ score of 58.3\% and  AP$_{25}$ score of 71.5\%, as shown in Table~\ref{tab:ab}. 
The whole MaskGroup model E achieves 41.9\%/62.7\%/73.6\% for AP/AP$_{50}$/AP$_{25}$ scores. It demonstrates that the mask branch can effectively refine the roughly clustered groups and is beneficial for the final predictions.
Meanwhile, compare to the baseline A, the model B encounters slight performance drop. This observation implies that it is necessary to use the mask branch together with the HPG module, because using the HPG without masking will bring some poorly grouped proposal instances.

\noindent\textbf{Without Hierarchical Grouping.}
We use one stage grouping strategy to cluster the points, which results in a degraded model C without using the hierarchical point grouping (HPG) algorithm. 
It achieves the AP score of 40.4\%,  AP$_{50}$ score of 59.8\%,  AP$_{25}$ score of 70.5\%, which are 1.5\%, 2.9\% and 3.1\% lower than the whole model with HPG (model E). It demonstrates that the proposed HPG algorithm is beneficial for 3D instance segmentation.

\noindent\textbf{Without MaskPool layer.}
We remove the MaskPool layer from our model to examine its effectiveness.
This degraded model D obtains the  AP score of 39.4\%,  AP$_{50}$ score of 61.1\%,  AP$_{25}$ score of 72.7\%, which are 2.5\%, 1.6\% and 0.9\% lower than the whole model E for AP/AP$_{50}$/AP$_{25}$ scores.  It demonstrates that the proposed MaskPool is beneficial for 3D instance segmentation.

\noindent\textbf{Impacts of Multi-scale Groups.}
By using hierarchical point grouping, we could make use of clusters in multi-scales for final predictions. Here we elaborate the effectiveness of this multi-scale strategy. 
Table~\ref{tab:ab_hgp} shows the results of using groups obtained with different clustering radii.
If we use only one clustering radius for grouping, the hierarchical grouping algorithm degrades into a common single-step algorithm.
A small grouping radius {$r$=0.01m} can mainly group small objects that are closely located.
It achieves the AP/AP$_{50}$/AP$_{25}$ scores of 37.5\%/59.4\%/71.0\%, which are not satisfactory, because it will drop many points in an object and makes the segmented instances incomplete.
The radius of {$r$=0.03} can achieve the best performance among all single radius settings.
With a larger radius {$r$=0.05}, the results begin to decrease as the large radius will cause some over-grouped instances.

Even though {$r$=0.01} can not perform well as other two larger radii, combining the groups obtained by  {$r$=0.01}  and  {$r$=0.03} takes advantage of both scales,
which results in better performance than single scale setting, \ie 62.6\% vs (59.4\%, 61.9\%) AP$_{50}$ scores.
Combining all multi-scale groups obtained with $r$=\{0.01, 0.03, 0.05\}, it achieves better performance.
With even more multi-scale groups, \ie, 4 to 6 scales, the performance will be further improved and begin to converge.
These results demonstrate that our hierarchical grouping algorithm can take advantage of multi-scale information to obtain better 3D instance segmentation results. Considering the trade-off between accuracy and complexity, we choose $r$=\{0.01, 0.03, 0.05\} in our model.

\section{Conclusion}
In this paper, we propose a novel framework named MaskGroup for accurate 3D instance segmentation.
To better group the points in 3D scenes, we propose a Hierarchical Point Grouping algorithm to merge them progressively into groups with different distances.
These multi-scale groups are then exploited for instance prediction, which is beneficial for predicting instances with different scales.
What's more, we propose a  MaskScoreNet to produce binary point masks for all grouped instances and effectively eliminate noisy points from the instances.
MaskGroup achieves 66.4\% AP$_{50}$ on the testing set of ScanNetV2 and outperforms prior methods, demonstrating the effectiveness of our proposed method.

{\small
\bibliographystyle{IEEEbib}
\bibliography{ref}

\begin{thebibliography}{10}

\bibitem{3dbonet}
Bo~Yang, Jianan Wang, Ronald Clark, Qingyong Hu, Sen Wang, Andrew Markham, and
  Niki Trigoni,
\newblock ``Learning object bounding boxes for 3d instance segmentation on
  point clouds,''
\newblock in {\em NeurIPS}, 2019.

\bibitem{3dsis}
Ji~Hou, Angela Dai, and Matthias Nie{\ss}ner,
\newblock ``3d-sis: 3d semantic instance segmentation of rgb-d scans,''
\newblock in {\em CVPR}, 2019.

\bibitem{jsis3d}
Quang{-}Hieu Pham, Duc~Thanh Nguyen, Binh{-}Son Hua, Gemma Roig, and Sai{-}Kit
  Yeung,
\newblock ``{JSIS3D:} joint semantic-instance segmentation of 3d point clouds
  with multi-task pointwise networks and multi-value conditional random
  fields,''
\newblock in {\em CVPR}, 2019.

\bibitem{asis}
Xinlong Wang, Shu Liu, Xiaoyong Shen, Chunhua Shen, and Jiaya Jia,
\newblock ``Associatively segmenting instances and semantics in point clouds,''
\newblock in {\em CVPR}, 2019.

\bibitem{3dmpa}
Francis Engelmann, Martin Bokeloh, Alireza Fathi, Bastian Leibe, and Matthias
  Nie{\ss}ner,
\newblock ``3d-mpa: Multi-proposal aggregation for 3d semantic instance
  segmentation,''
\newblock in {\em CVPR}, 2020.

\bibitem{pointgroup}
Li~Jiang, Hengshuang Zhao, Shaoshuai Shi, Shu Liu, Chi{-}Wing Fu, and Jiaya
  Jia,
\newblock ``Pointgroup: Dual-set point grouping for 3d instance segmentation,''
\newblock in {\em CVPR}, 2020.

\bibitem{occseg}
Lei Han, Tian Zheng, Lan Xu, and Lu~Fang,
\newblock ``Occuseg: Occupancy-aware 3d instance segmentation,''
\newblock in {\em CVPR}, 2020.

\bibitem{liang2021instance}
Zhihao Liang, Zhihao Li, Songcen Xu, Mingkui Tan, and Kui Jia,
\newblock ``Instance segmentation in 3d scenes using semantic superpoint tree
  networks,''
\newblock in {\em ICCV}, 2021, pp. 2783--2792.

\bibitem{chen2021hierarchical}
Shaoyu Chen, Jiemin Fang, Qian Zhang, Wenyu Liu, and Xinggang Wang,
\newblock ``Hierarchical aggregation for 3d instance segmentation,''
\newblock in {\em ICCV}, 2021, pp. 15467--15476.

\bibitem{3drcnn}
Abhijit Kundu, Yin Li, and James~M. Rehg,
\newblock ``3d-rcnn: Instance-level 3d object reconstruction via
  render-and-compare,''
\newblock in {\em CVPR}, 2018.

\bibitem{sgpn}
Weiyue Wang, Ronald Yu, Qiangui Huang, and Ulrich Neumann,
\newblock ``{SGPN:} similarity group proposal network for 3d point cloud
  instance segmentation,''
\newblock in {\em CVPR}, 2018.

\bibitem{votenet}
Charles~R. Qi, Or~Litany, Kaiming He, and Leonidas~J. Guibas,
\newblock ``Deep hough voting for 3d object detection in point clouds,''
\newblock in {\em ICCV}, 2019.

\bibitem{shi2019hierarchy}
Yifei Shi, Angel~X Chang, Zhelun Wu, Manolis Savva, and Kai Xu,
\newblock ``Hierarchy denoising recursive autoencoders for 3d scene layout
  prediction,''
\newblock in {\em CVPR}, 2019, pp. 1771--1780.

\bibitem{armeni20193d}
Iro Armeni, Zhi-Yang He, JunYoung Gwak, Amir~R Zamir, Martin Fischer, Jitendra
  Malik, and Silvio Savarese,
\newblock ``3d scene graph: A structure for unified semantics, 3d space, and
  camera,''
\newblock in {\em ICCV}, 2019, pp. 5664--5673.

\bibitem{sparse3d}
Benjamin Graham, Martin Engelcke, and Laurens van~der Maaten,
\newblock ``3d semantic segmentation with submanifold sparse convolutional
  networks,''
\newblock in {\em CVPR}, 2018.

\bibitem{mtml}
Jean Lahoud, Bernard Ghanem, Marc Pollefeys, and Martin~R Oswald,
\newblock ``3d instance segmentation via multi-task metric learning,''
\newblock in {\em ICCV}, 2019.

\bibitem{fpn}
Tsung{-}Yi Lin, Piotr Doll{\'{a}}r, Ross~B. Girshick, Kaiming He, Bharath
  Hariharan, and Serge~J. Belongie,
\newblock ``Feature pyramid networks for object detection,''
\newblock in {\em CVPR}, 2017.

\bibitem{chen2021empirical}
Xinghao Chen, Chang Xu, Minjing Dong, Chunjing Xu, and Yunhe Wang,
\newblock ``An empirical study of adder neural networks for object detection,''
\newblock {\em NeurIPS}, vol. 34, 2021.

\bibitem{DBLP:conf/iccv/ArmeniHZGMFS19}
Iro Armeni, Zhi{-}Yang He, Amir~Roshan Zamir, JunYoung Gwak, Jitendra Malik,
  Martin Fischer, and Silvio Savarese,
\newblock ``3d scene graph: {A} structure for unified semantics, 3d space, and
  camera,''
\newblock in {\em ICCV}, 2019.

\bibitem{DBLP:conf/cvpr/ShiCWS019}
Yifei Shi, Angel~X. Chang, Zhelun Wu, Manolis Savva, and Kai Xu,
\newblock ``Hierarchy denoising recursive autoencoders for 3d scene layout
  prediction,''
\newblock in {\em CVPR}, 2019.

\bibitem{jiang2018acquisition}
Borui Jiang, Ruixuan Luo, Jiayuan Mao, Tete Xiao, and Yuning Jiang,
\newblock ``Acquisition of localization confidence for accurate object
  detection,''
\newblock in {\em ECCV}, 2018.

\bibitem{li2019gs3d}
Buyu Li, Wanli Ouyang, Lu~Sheng, Xingyu Zeng, and Xiaogang Wang,
\newblock ``Gs3d: An efficient 3d object detection framework for autonomous
  driving,''
\newblock in {\em CVPR}, 2019.

\bibitem{saloss}
Zhidong Liang, Ming Yang, Hao Li, and Chunxiang Wang,
\newblock ``3d instance embedding learning with a structure-aware loss function
  for point cloud segmentation,''
\newblock {\em IEEE RA-L}, 2020.

\bibitem{pfusion}
Gaku Narita, Takashi Seno, Tomoya Ishikawa, and Yohsuke Kaji,
\newblock ``Panopticfusion: Online volumetric semantic mapping at the level of
  stuff and things,''
\newblock in {\em IROS}, 2019.

\bibitem{gicn}
Shih{-}Hung Liu, Shang{-}Yi Yu, Shao{-}Chi Wu, Hwann{-}Tzong Chen, and
  Tyng{-}Luh Liu,
\newblock ``Learning gaussian instance segmentation in point clouds,''
\newblock {\em CoRR}, 2020.

\bibitem{dyco}
Tong He, Chunhua Shen, and Anton van~den Hengel,
\newblock ``{DyCo3d}: Robust instance segmentation of 3d point clouds through
  dynamic convolution,''
\newblock in {\em CVPR}, 2021.

\bibitem{pe}
Biao Zhang and Peter Wonka,
\newblock ``Point cloud instance segmentation using probabilistic embeddings,''
\newblock in {\em CVPR}, 2021.

\bibitem{s3dis}
Iro Armeni, Ozan Sener, Amir~R Zamir, Helen Jiang, Ioannis Brilakis, Martin
  Fischer, and Silvio Savarese,
\newblock ``3d semantic parsing of large-scale indoor spaces,''
\newblock in {\em CVPR}, 2016.

\bibitem{he2020learning}
Tong He, Dong Gong, Zhi Tian, and Chunhua Shen,
\newblock ``Learning and memorizing representative prototypes for 3d point
  cloud semantic and instance segmentation,''
\newblock in {\em ECCV}, 2020.

\bibitem{he2020instance}
Tong He, Yifan Liu, Chunhua Shen, Xinlong Wang, and Changming Sun,
\newblock ``Instance-aware embedding for point cloud instance segmentation,''
\newblock in {\em ECCV}, 2020.

\bibitem{chu2021icm}
Ruihang Chu, Yukang Chen, Tao Kong, Lu~Qi, and Lei Li,
\newblock ``Icm-3d: Instantiated category modeling for 3d instance
  segmentation,''
\newblock {\em IEEE RA-L}, 2021.

\bibitem{scannet}
Angela Dai, Angel~X Chang, Manolis Savva, Maciej Halber, Thomas Funkhouser, and
  Matthias Nie{\ss}ner,
\newblock ``Scannet: Richly-annotated 3d reconstructions of indoor scenes,''
\newblock in {\em CVPR}, 2017.

\end{thebibliography}
}

\clearpage

\onecolumn
\section{Appendix}

\begin{table*}[htb]
	\vspace{-5mm}
	\centering
	\resizebox{\textwidth}{!}{
		\begin{tabular}{l|c|cccccccccccccccccc}
			\toprule %
			Method & Avg AP$_{50}$ &\rotatebox{90}{bathtub}&\rotatebox{90}{bed}&\rotatebox{90}{bookshe.}&\rotatebox{90}{cabinet}&\rotatebox{90}{chair}&\rotatebox{90}{counter}&\rotatebox{90}{curtain}&\rotatebox{90}{desk}&\rotatebox{90}{door}&\rotatebox{90}{otherfu.}&\rotatebox{90}{picture}&\rotatebox{90}{refrige.}&\rotatebox{90}{s. curtain}&\rotatebox{90}{sink}&\rotatebox{90}{sofa}&\rotatebox{90}{table}&\rotatebox{90}{toilet}&\rotatebox{90}{window}\\
			\midrule
			SGPN~\cite{sgpn} & 0.143 & 0.208 & 0.390 & 0.169 & 0.065 & 0.275 & 0.029 & 0.069 & 0.000 & 0.087 & 0.043 & 0.014 & 0.027 & 0.000 & 0.112 & 0.351 & 0.168 & 0.438 & 0.138 \\
			3D-SIS~\cite{3dsis} & 0.382 & 1.000 & 0.432 & 0.245 & 0.190 & 0.577 & 0.013 & 0.263 & 0.033 & 0.320 & 0.240 & 0.075 & 0.422 & 0.857 & 0.117 & 0.699 & 0.271 & 0.883 & 0.235 \\
			PanopticFusion~\cite{pfusion} & 0.478 & 0.667 & 0.712 & 0.595 & 0.259 & 0.550 & 0.000 & 0.613 & 0.175 & 0.250 & 0.434 & 0.437 & 0.411 & 0.857 & 0.485 & 0.591 & 0.267 & 0.944 & 0.35 \\
			3D-BoNet~\cite{3dbonet} & 0.488 & \textbf{1.000} & 0.672 & 0.590 & 0.301 & 0.484 & 0.098 & 0.620 & 0.306 & 0.341 & 0.259 & 0.125 & 0.434 & 0.796 & 0.402 & 0.499 & 	0.513 & 0.909 & 0.439 \\
			MTML~\cite{mtml} & 0.549 & \textbf{1.000} & {0.807} & 0.588 & 0.327 & 0.647 & 0.004 & {0.815} & 0.180 & 0.418 & 0.364 & 0.182 & 0.445 & 1.000 & 0.442 & 0.688 & {0.571} & \textbf{1.000} & 0.396 \\ %
			3D-MPA~\cite{3dmpa} &0.611 &\textbf{1.000} &0.833 &0.765 &0.526 &0.756 &0.136 &0.588 &0.470 &0.438 &0.432 &0.358 &0.650 &0.857 &0.429 &0.765 &0.557 &\textbf{1.000} &0.430 \\
			{PointGroup}~\cite{pointgroup}  & {0.636} & \textbf{1.000} & 0.765 & {0.624} & {0.505} & {0.797} & 0.116 & 0.696 & {0.384} & {0.441} & {0.559} & {0.476} & {0.596} & \textbf{1.000} & {0.666} & {0.756} & 0.556 & 0.997 & {0.513} \\
			{GICN}~\cite{gicn}    &0.638 &\textbf{1.000} &\textbf{0.895} &\textbf{0.800} &0.480 &0.676 &\textbf{0.144} &{0.737} &0.354 &0.447 &0.400 &0.365 &\textbf{0.700} &\textbf{1.000} &0.569 &\textbf{0.836} &0.599 &\textbf{1.000} &0.473    \\
			
			{DyCo3D}~\cite{dyco} &0.641 &\textbf{1.000} &0.841 &0.893 &0.531 &0.802 &0.115 &0.588 &0.448 &0.438 &0.537 &0.430 &0.550 &0.857 &0.534 &0.764 &\textbf{0.657} &0.987 &0.568 \\
			
			{PE}~\cite{pe} &0.645 &\textbf{1.000} &0.773 &0.798 &0.538 &0.786 &0.088 &\textbf{0.799} &0.350 &0.435 &0.547 &0.545 &0.646 &0.933 &0.562 &0.761 &0.556 &0.997 &0.501 \\
			
			\textbf{MaskGroup (Ours)} &\textbf{0.664} &\textbf{1.000} &0.822 &0.764 &\textbf{0.616} &\textbf{0.815} &0.139 &0.694 &\textbf{0.597} &\textbf{0.459} &\textbf{0.566} &\textbf{0.599} &0.600 &0.516 &\textbf{0.715} &0.819 &{0.635} &\textbf{1.000} &\textbf{0.603} \\
			\bottomrule %
		\end{tabular}
	}
	\vspace{-2mm}
	\caption{3D instance segmentation results on ScanNetV2~\cite{scannet} testing set with AP$_{50}$ scores. Our method yields the highest average AP$_{50}$ performance among all existing methods published in the literature.}
	\vspace{-1mm}
	\label{tab:scannet-test}
\end{table*}

\begin{figure*}[htb]
	\begin{center}
		\includegraphics[width=0.72\linewidth]{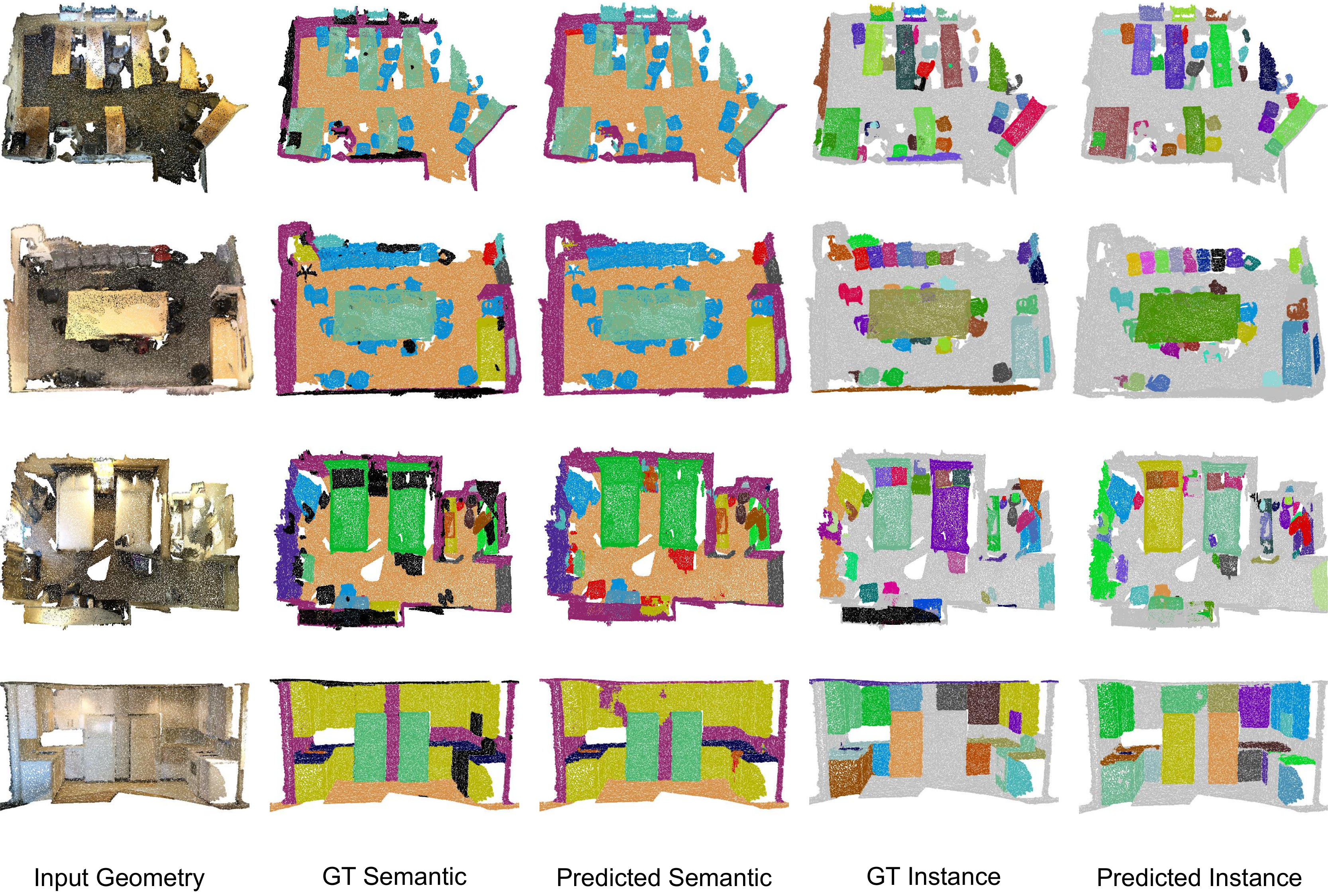}
	\end{center}
	\vspace{-4.5mm}
	\caption{Some 3D instance segmentation results on the validation set of ScanNetV2 obtained by our method.}
	\label{fig:results}
	\vspace{-2.mm}
\end{figure*}

\begin{figure*}[htb]
	\begin{center}
		\includegraphics[width=0.85\linewidth]{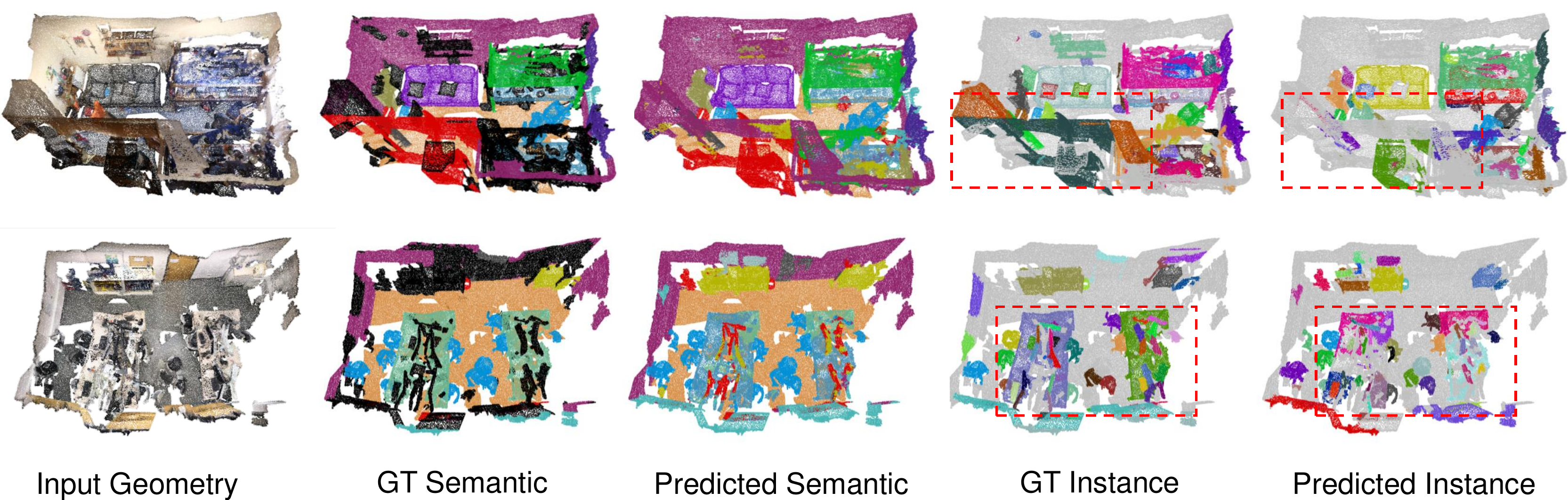}
	\end{center}
	\vspace{-1.5em}
	\caption{{Visualization of some failure cases.}}
	\label{fig:bad_case}
\end{figure*}

\begin{figure*}[htb]
	\begin{center}
		\includegraphics[width=0.78\linewidth]{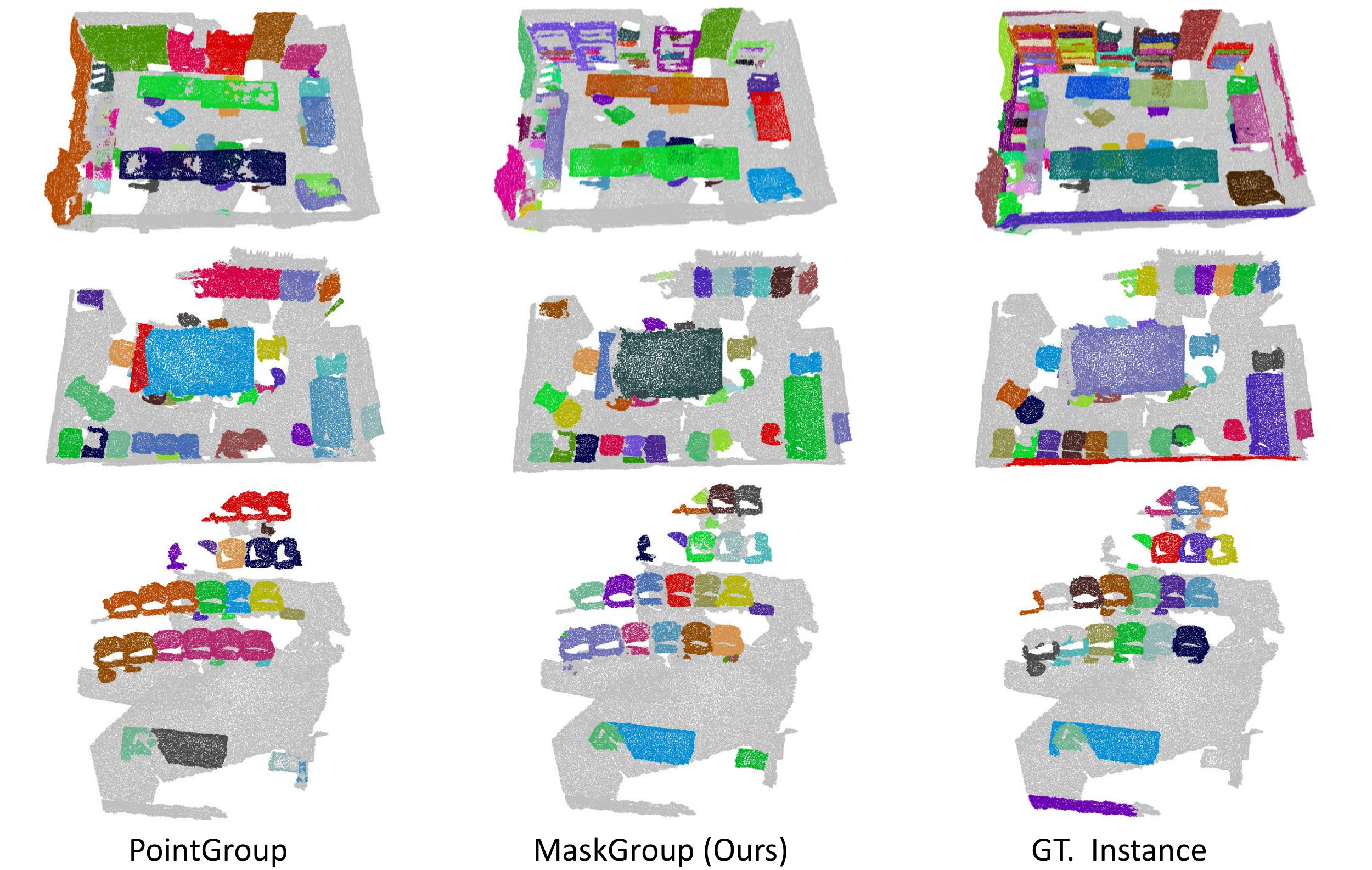}
	\end{center}
	\vspace{-1.2em}
	\caption{Comparisons of PointGroup and our proposed MaskGroup.}
	\label{fig:sota}
\end{figure*}

\begin{figure*}[htb]
	\begin{center}
		\includegraphics[width=0.85\linewidth]{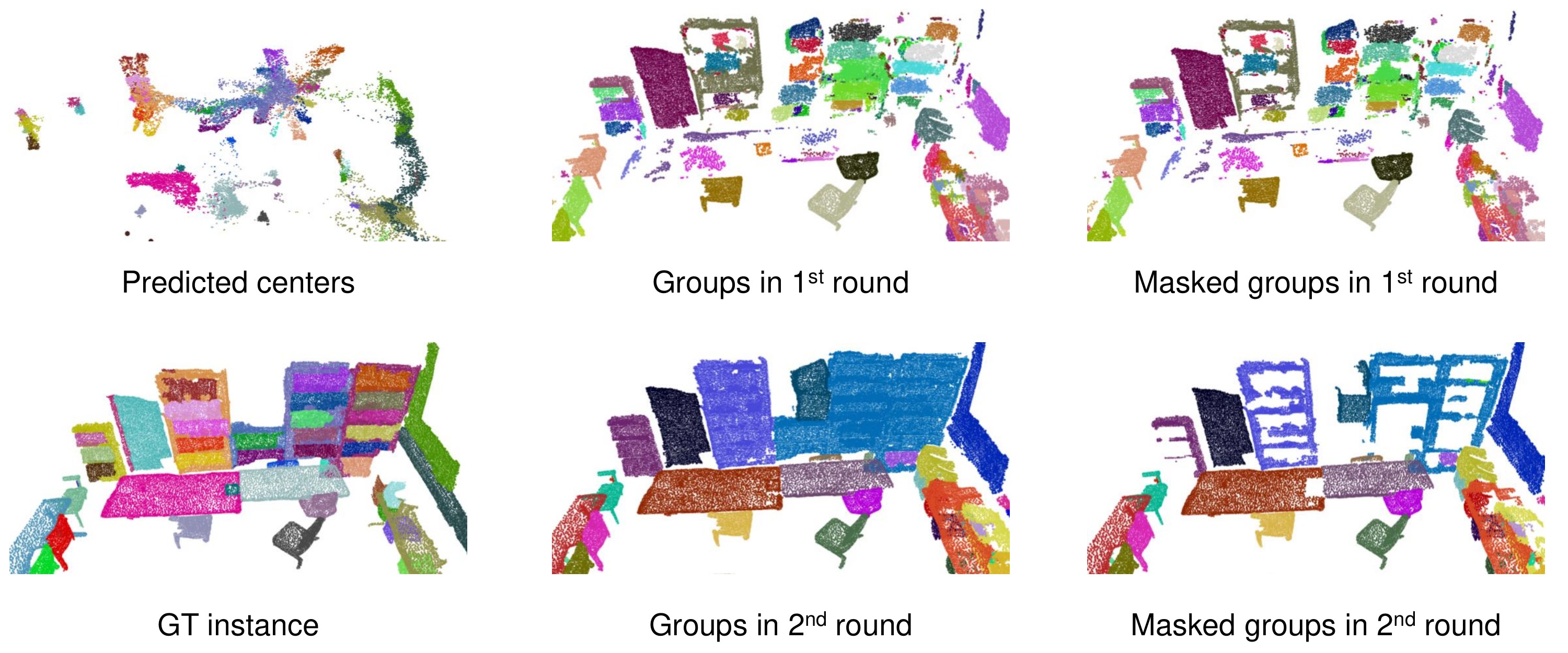}
	\end{center}
	\vspace{-1em}
	\caption{Visualization of the hierarchical point grouping and the mask prediction.}
	\label{fig:ablation}
\end{figure*}

Table~\ref{tab:scannet-test} lists detailed performance for each category on ScanNetV2 benchmark. Our proposed MaskGroup ranks the $1^{st}$ place in 10 out of 18 classes.

Fig.~\ref{fig:results} shows the qualitative results on the validation set of ScanNetV2.
The scene in the first row mainly contains some tables and chairs, which can well be separated using the void space.
The second row shows a scene with some chairs which are lie very close spatially and our model segments these chairs well.
We can also segment different ``pillow'', even they are located on the bed and are difficult to recognize as shown in the third row. 
Some instances are not annotated in the ground-truth, like the ``refrigerator'' in the last row. However, our method can also produce good segmentation results.
These results demonstrate that the proposed approach achieves robust instance segmentation results for complex environments.

We provide some qualitative results of PointGroup~\cite{pointgroup} and our proposed MaskGroup, as shown in Fig.~\ref{fig:sota}. Our method obtains better results than PointGroup.
We also visualize several failure cases in Fig.~\ref{fig:bad_case}. In the first row, the doors and cabinets are not perfectly segmented, which exhibit quite similar geometric shapes and close spatial locations. In the second row, there are massive of small objects in the table, making it quite challenging to achieve accuracy instance segmentation.

Fig.~\ref{fig:ablation} shows the qualitative results of the effectiveness of the mask branch. 
As we can see, the mask prediction branch can mask out some wrong points for the $2^{nd}$ round groups.
For example, the instance ``bookshelf'' has a complicated shape, which is difficult for point grouping. With the help of the proposed mask branch, the instance of the ``bookshelf'' is well segmented after the mask operation.

\end{document}